\documentclass[10pt, a4paper]{article}
\usepackage{lrec}
\usepackage{multibib}
\usepackage{graphicx}
\usepackage{tabularx}
\usepackage{soul}

\usepackage{booktabs}

\usepackage{epstopdf}
\usepackage[latin1]{inputenc}

\usepackage{hyperref}
\usepackage{xstring}
\makeatletter
\renewcommand{\p@section}{\arabic{section}\expandafter\@gobble}
\renewcommand{\p@subsection}{\thesection\arabic{subsection}\expandafter\@gobble
}
\renewcommand{\p@subsubsection}{\thesubsection\arabic{subsubsection}\expandafter\@gobble
}
\makeatother

\title{A Resource for Studying Chatino Verbal Morphology}

\name{Hilaria Cruz\textsuperscript{1}, Antonios Anastasopoulos\textsuperscript{2}, and Gregory Stump\textsuperscript{3}}

\address{\textsuperscript{1}University of Louisville, 2211 South Brook, Louisville KY 40292, USA,\\
\textsuperscript{2}Carnegie Mellon University, 5000 Forbes Ave, Pittsburgh PA 15213, USA,\\ \textsuperscript{3}University of Kentucky, 13964 West 144th Court, Olathe KS 66062, USA \\
         hilaria.cruz@louisville.edu, aanastas@cs.cmu.edu, gregorystump76@gmail.com\\}

\abstract{
We present the first resource focusing on the verbal inflectional morphology of San Juan Quiahije Chatino, a tonal mesoamerican language spoken in Mexico. We provide a collection of complete inflection tables of~198 lemmata, with morphological tags based on the UniMorph schema. We also provide baseline results on three core NLP tasks: morphological analysis, lemmatization, and morphological inflection.\\ 
\newline  }

\begin{document}

\maketitleabstract

\section{Introduction}

The recent years have seen unprecedented forward steps for Natural Language Processing (NLP) over almost every NLP subtask, relying on the advent of large data collections that can be leveraged to train deep neural networks.
However, this progress has solely been observed in languages with significant data resources, while low-resource languages are left behind.

The situation for endangered languages is usually even worse, as the focus of the scientific community mostly relies in \textit{language documentation}.
The typical endangered language documentation process typically includes the creation of language resources in the form of word lists, audio and video recordings, notes, or grammar fragments, with the created resources then stored into  large online linguistics archives.
This process is often hindered by the so-called Transcription Bottleneck, but recent advances \cite{cavar2016endangered,adams2018evaluation} provide promising directions towards a solution for this issue.

However, language documentation and linguistic description, although extremely important itself, does not meaningfully contribute to \textit{language conservation}, which aims to ensure that the language stays in use. We believe that a major avenue towards continual language use is by further creating language technologies for endangered languages, essentially elevating them to the same level as high-resource, economically or politically stronger languages.

The majority of the world's languages are categorized as synthetic, meaning that they have rich morphology, be it fusional, agglutinative, polysynthetic, or a mixture thereof.
As Natural Language Processing (NLP) keeps expanding its frontiers to encompass more and more languages, modeling of the grammatical functions that guide language generation is of utmost importance.
It follows, then, that the next crucial step for expanding NLP research on endangered languages is creating benchmarks for standard NLP tasks in such languages.

With this work we take a small first step towards this direction. We present a resource that allows for benchmarking two NLP tasks in San Juan Quiahije Chatino, an endangered language spoken in southern Mexico: morphological analysis and morphological inflection, with a focus on the verb morphology of the language.

We first briefly discuss the Chatino language and the intricacies of its verb morphology (\S\ref{sec:chatino}), then describe the resource (\S\ref{sec:resource}), and finally present baseline results on both the morphological analysis and the inflection tasks using state-of-the-art neural models (\S\ref{sec:baselines}). 
We make our resource publicly available online\footnote{\url{https://github.com/antonisa/chatino_inflection_paradigms}}.

\section{The Chatino Language}
\label{sec:chatino}
Chatino is a group of languages spoken in Oaxaca, Mexico. Together with the Zapotec language group, the Chatino languages form the Zapotecan branch of the Otomanguean language family.
There are three main Chatino languages: Zenzontepec Chatino
(ZEN, ISO 639-2 code czn), Tataltepec Chatino (TAT, cta), and Eastern Chatino (ISO 639-2 ctp, cya, ctz, and cly) (E.Cruz 2011 and Campbell 2011). San Juan Quiahije
Chatino (SJQ), the language of the focus of this study, belongs to Eastern Chatino, and is used by about~3000 speakers.

\begin{table}[t]
    \centering
    \begin{tabular}{c|cccc}
    \toprule
    & All & Train & Dev & Test \\
        Paradigms & 198  \\
        Verb Classes & 29 \\
        Forms & 4716 & 3774 & 471 & 471 \\
    \bottomrule
    \end{tabular}
    \caption{Basic Statistics of our resource.}
    \label{tab:stats}
\end{table}

\paragraph{Typology and Writing System}
Eastern Chatino languages , including SJQ Chatino, are intensively tonal~\cite{cruz2004phonological,cruz2014finding}. Tones mark both lexical and grammatical
distinctions in Eastern Chatino languages. 

In SJQ Chatino, there are eleven tones. Three different systems for representing tone distinctions are employed in the literature: the S-H-M-L system of~\cite{cruz2004phonological}; the numeral system of~\cite{cruz2014linguistic}; and the alphabetic system of~\cite{cruz2014finding}. The correspondences among these three systems are given in Table~\ref{tab:writing}. 
For present purposes, we will use numeral representations of the second sort. The number 1 represents a high pitch, 4 represents a low pitch, and double digits represent contour tones.

\begin{table}[t]
    \centering
    \begin{tabular}{c|ccc}
    \toprule
Tone & S-H-M-L & Numeral & Alphabetic \\ 
description & \tiny{(Cruz 2011)} & \tiny{(Cruz 2014)} & \tiny{(Cruz \& Woodbury 2013)} \\ 
 \midrule
high & H  & 1 & E \\ 
high-superhigh & HS & 10 & D \\ 
high-low & HL & 14 & B \\ 
mid & M  & 2 & C \\ 
mid-superhigh & MS & 20 & H \\ 
mid-high & MH & 32 & I \\ 
mid-low & ML  & 24 & J \\ 
low & L  & 4 & A \\ 
low-superhigh & LS  & 40 & M  \\ 
low-high & LH & 42 & G \\ 
low-mid & LM & 3 & F \\ 
\bottomrule
    \end{tabular}
    \caption{Three alternative systems for representing 
the SJQ Chatino tones.}
    \label{tab:writing}
\end{table}

\paragraph{Verb Morphology}

SJQ Chatino verb inflection distinguishes four aspect/mood categories: completive (\textit{`I did'}), progressive (\textit{`I am doing'}), habitual (\textit{`I habitually do'}) and potential (\textit{`I might do'}). In each of these categories, verbs inflect for three persons (first, second, third) and two numbers (singular, plural) and distinguish inclusive and exclusive categories of the first person plural (\textit{`we including you'} vs \textit{`we excluding you'}). Verbs can be classified into dozens of different conjugation classes. Each conjugation class involves its own tone pattern; each tone pattern is based on a series of three person/number (PN) triplets. A PN triplet [X, Y, Z] consists of three tones: tone X is employed in the third person singular as well as in all plural forms; tone Y is employed in the second person singular, and tone Z, in the third person singular. Thus, a verb's membership in a particular conjugation class entails the assignment of one tone triplet to completive forms, another to progressive forms, and a third to habitual and potential forms. The paradigm of the verb lyu1 \textit{`fall'} in Table~\ref{tab:example} illustrates: the conjugation class to which this verb belongs entails the assignment of the triplet [1, 42, 20] to the completive, [1, 42, 32] to the progressive, and [20, 42, 32] to the habitual and potential. Verbs in other conjugation classes exhibit other triplet series.\footnote{A more thorough introduction into Chatino verbal morphology will appear at \cite{cruzstump2020}.}

\begin{table*}[t]
    \centering
    \begin{tabular}{cc|cccc}
    \toprule
&\multicolumn{1}{c}{Aspect:} & CPL & PROG & HAB & POT \\ 
\midrule
\multicolumn{6}{l}{ndyu2 `fell from above'} \\ 
&PN triple: & 2-1-40 & 2-1-40 & \multicolumn{2}{c}{2-1-40}  \\
\midrule
Singular&1 & ndyon40 & ndyon40 & ndyon40 & tyon40 \\ 
&2 & ndyu1 & ndyu1 & ndyu1 & tyu1 \\ 
&3 & ndyu2 & ndyu2 & ndyu2 & tyu2 \\ 
Plural&1 inclusive & ndyon2on1 & ndyon2on1 & ndyon2on1 & ntyon2on1 \\ 
&1 exclusive & ndyu2 wa42 & ndyu2 wa42 & ndyu2 wa42 & ntyu2 wa42 \\ 
&2 & ndyu2 wan1 & ndyu2 wan1 & ndyu2 wan1 & ntyu2 wan1 \\ 
&3 & ndyu2 renq1 & ndyu2 renq1 & ndyu2 renq1 & ntyu2 renq1 \\
\midrule
\multicolumn{6}{l}{lyu1 `to fall'} \\ 
&PN triple: & 1-42-20 & 1-42-32 & \multicolumn{2}{c}{20-42-32} \\
\midrule
Singular&1 & lyon20 & nlyon32 & nlyon32 & klyon32 \\ 
&2 & lyu42 & nlyu42 & nlyu42 & klyu42 \\ 
&3 & lyu1 & nlyu1 & nlyu20 & klyu20 \\ 
Plural&1 inclusive & lyon1on1 & nlyon1on1 & nlyon20on32 & klyon20on32 \\ 
&1 exclusive & lyu1 wa42 & nlyu1 wa42 & nlyu20 wa42 & klyu20 wa42 \\ 
&2 & lyu1 wan24 & lyu1 wan24 & nlyu20 wan24 & klyu20 wan24 \\ 
&3 & lyu1 renq24 & lyu1 renq24 & nlyu20 renq24 & klyu20 renq24 \\
\bottomrule
\end{tabular}
    \caption{Complete inflection paradigms for two lemmata: one with a single PN triple across all aspects (top), and one with three different PN triples (bottom).}
    \label{tab:example}
\end{table*}

\section{The Resource}
\label{sec:resource}

We provide a hand-curated collection of complete inflection tables for 198 lemmata. The morphological tags follow the guidelines of the UniMorph schema \cite{sylak2016composition,kirov2018unimorph}, in order to allow for the potential of cross-lingual transfer learning, and they are tagged with respect to:
\begin{itemize}
    \item Person: first (1), second (2), and third (3)
    \item Number: singular (SG) ad plural (PL)
    \item Inclusivity (only applicable to first person plural verbs: inclusive (INCL) and exclusive (EXCL)
    \item Aspect/mood: completive (CPL), progressive (PROG), potential (POT), and habitual (HAB).
\end{itemize}

Two examples of complete inflection tables for the verbs \textit{ndyu2} `fell from above' and \textit{lyu1} `fall' are shown in Table~\ref{tab:example}. Note how the first verb has the same PN triplet for all four aspect/mood categories, while the second paradigm is more representative in that it involves three different triplets (one for the completive, another for the progressive, and another for the habitual/potential).
This variety is at the core of why the SJQ verb morphology is particularly interesting, and a challenging testcase for modern NLP systems.
 
In total, we end up with 4716 groupings (triplets) of a lemma, a tag-set, and a form; we split these groupings randomly into a training set (3774 groupings), a development set (471 groupings), and test set (471 groupings). Basic statistics of the corpus are outlined in Table 1~\ref{tab:stats}. Compared to all the other languages from the Unimorph project, this puts SJQ Chatino in the low- to mid-resource category, but nonetheless it is more than enough for benchmarking purposes.\footnote{In future work we will investigate whether more controlled training-development-test splits such that the splits is non-random but rather across whole lemmata or even across whole verb classes results in different generalization issues.}


\section{Baseline Results}
\label{sec:baselines}

\paragraph{Inflectional realization}
\begin{table}[t]
    \centering
    \begin{tabular}{c|cc}
    \toprule
        Setting & Accuracy & Levenshtein distance\\
    \midrule
        standard & 60\% & 0.92 \\
        +hallucinated data & 62\% & 1.02\\
    \bottomrule
    \end{tabular}
    \caption{Morphological Inflection Results}
    \label{tab:infl_res}
\end{table}
Inflectional realization defines the inflected forms of a lexeme/lemma. As a computational task, often referred to as simply ``morphological inflection," inflectional realization is framed as a mapping from the pairing of a lemma with a set of morphological tags to the corresponding word form. For example, the inflectional realization of SJQ Chatino verb forms entails a mapping of the pairing of the lemma lyu1 `fall' with the tag-set {1;SG;PROG} to the word form nlyon32.


Morphological inflection has been thoroughly studied in monolingual high resource settings, especially through the recent SIGMORPHON challenges \cite{cotterell2016sigmorphon,cotterell-etal-2017-conll,cotterell-etal-2018-conll}, with the latest iteration focusing more on low-resource settings, utilizing cross-lingual transfer~\cite{mccarthy2019sigmorphon}.
We use the guidelines of the state-of-the-art approach of \cite{anastasopoulos-neubig-2019-pushing} that achieved the highest inflection accuracy in the latest SIGMORPHON 2019 morphological inflection shared task.
Our models are implemented in DyNet~\cite{neubig2017dynet}.

Inflection results are outlined in Table~\ref{tab:infl_res}. 
In the `standard' setting we simply train on the pre-defined training set, achieving an exact-match accuracy of~60\% over the test set.
Interestingly, the data augmentation approach of~\cite{anastasopoulos-neubig-2019-pushing} that hallucinates new training paradigms based on character level alignments does not heed significant improvements in accuracy (only~2 percentage points increase, cf. with more than~15 percentage points increases in other languages).
These results indicate that automatic morphological inflection for low-resource tonal languages like SJQ Chatino poses a particularly challenging setting, which perhaps requires explicit handling of tone information by the model.

\paragraph{Morphological Analysis}
\begin{table}[t]
    \centering
    \begin{tabular}{c|c}
    \toprule
        Setting & Exact Match Accuracy\\
    \midrule
        standard & 67\%\\
    \bottomrule
    \end{tabular}
    \caption{Morphological Analysis Results}
    \label{tab:morph_res}
\end{table}
Morphological analysis is the task of creating a morphosyntactic description for a given word. It can be framed in a context-agnostic manner (as in our case) or within a given context, as for instance for the SIGMORPHON 2019 second shared task~\cite{mccarthy2019sigmorphon}.
We trained an encoder-decoder model that receives the form as character-level input, encodes it with a BiLSTM encoder, and then an attention enhanced decoder~\cite{bahdanau2014neural} outputs the corresponding sequence of morphological tags, implemented in DyNet. The baseline results are shown in Table~\ref{tab:morph_res}. The exact-match accuracy of~67\% is lower than the average accuracy that context-aware systems can achieve, and it highlights the challenge that the complexity of the tonal system of SJQ Chatino can pose.

\paragraph{Lemmatization}
\begin{table}[t]
    \centering
    \begin{tabular}{c|cc}
    \toprule
        Input & Accuracy & Levenshtein distance\\
    \midrule
        form (no tags) & 89\% & 0.27 \\
        form + tags & 90\% & 0.21\\
    \bottomrule
    \end{tabular}
    \caption{Lemmatization Results.}
    \label{tab:lemma_res}
\end{table}
Lemmatization is the task of retrieving the underlying lemma from which an inflected form was derived.
Although in some languages the lemma is distinct from all forms, in SJQ Chatino the lemma is defined as the completive third-person singular form.
As a computational task, lemmatization entails producing the lemma given an inflected form (and possibly, given a set of morphological tags describing the input form). Popular approaches tackle it as a character-level edit sequence generation task~\cite{chrupala2006simple}, or as a character-level sequence-to-sequence task~\cite{bergmanis-goldwater-2018-context}.
For our baseline lemmatization systems we follow the latter approach. We trained a character level encoder-decoder model, similar to the above-mentioned inflection system, implemented in DyNet.

The baseline results, with and without providing gold morphological tags along with the inflected form as input, are outlined in Table~\ref{tab:lemma_res}. We find that automatic lemmatization in SJQ Chatino achieves fairly high accuracy even with our simple baseline models (89\% accuracy, $0.27$ average Levenshtein distance) and that providing the gold morphological tags provides a performance boost indicated by small improvements on both metrics.
It it worth noting, though, that these results are also well below the $94--95\%$ average accuracy and $0.13$ average Levenshtein distance that lemmatization models achieved over~107 treebanks in~66 languages for the SIGMORPHON 2019 shared task~\cite{mccarthy2019sigmorphon}.

\section{Related Work}
Our work builds and expands upon previous work on Indigenous languages of the Americas specifically focusing on the complexity of their morphology.
Among other works similar to ours, \cite{cox2016computational} focused on the morphology of Dene verbs, \cite{moeller2018neural} on Arapaho verbs, \cite{cardenas2018morphological} on Shipibo-Konibo, and \cite{chen-schwartz-2018-morphological} on Saint Lawrence Island and Central Siberian Yupik.
\cite{sylak2016remote} describe an approach for elicit complete inflection paradigms, with experiments in languages like Nahuatl.
Our resource is the first one for SJQ Chatino, but it also provides an exciting new data point in the computational study of morphological analysis, lemmatization, and inflection, as it is the first one in a tonal language with explicit tonal markings in the writing system.
In a similar vein, the Oto-Manguean Inflectional Class Database\footnote{\url{http://www.oto-manguean.surrey.ac.uk/}} \cite{palancar2015oto} provides a valuable resource for studying the verbal morphology of Oto-Manguean languages (including a couple of other Chatino variants: Yaitepec and Zenzotepec Chatino) but not in a format suitable for computational experiments.

\section{Conclusion}
We presented a resource of~198 complete inflectional paradigms in San Juan Quiahije Chatino, which will facilitate research in computational morphological analysis and inflection for low-resource tonal languages and languages of Mesoamerica. We also provide strong baseline results on computational morphological analysis, lemmatization, and inflection realization, using character-level neural encoder-decoder systems.

For future work, while we will keep expanding our resource to include more paradigms, we will also follow the community guidelines in extending our resource to include morphological analysis and inflection examples \textit{in context}.

\section{Acknowledgements}
Part of this work was done during the Workshop on Language Technology for Language Documentation and Revitalization.\footnote{\url{https://sites.google.com/view/ltldr/home}}
This material is based upon work generously supported by the National Science Foundation under grant 1761548.

\section{Bibliographical References}
\label{main:ref}
\bibliographystyle{lrec}
\bibliography{References}


\end{document}